\documentclass[lettersize,journal]{IEEEtran}
\usepackage{amsmath,amsfonts}
\usepackage{algorithmic}
\usepackage{algorithm}
\usepackage{array}
\usepackage[caption=false,font=normalsize,labelfont=sf,textfont=sf]{subfig}
\usepackage{textcomp}
\usepackage{stfloats}
\usepackage{url}
\usepackage{verbatim}
\usepackage{graphicx}
\usepackage{cite}
\usepackage{graphicx}
\usepackage{algorithmic}
\usepackage{algorithm}
\usepackage{amsmath}
\usepackage{amssymb}
\usepackage{makecell}
\usepackage{hyperref}
\usepackage{mathtools}

\usepackage{xcolor}
\usepackage{subcaption}

\usepackage[inkscapelatex=false]{svg}

\usepackage{adjustbox} 

\usepackage{diagbox}
\usepackage{multirow}
\usepackage{makecell} 

\usepackage{bbm}

\usepackage[table,xcdraw]{xcolor}

\definecolor{cgreen}{RGB}{0, 255, 0}
\definecolor{xred}{RGB}{255, 0, 0}

\usepackage{pifont}
\colorlet{blue}{black}
\colorlet{magenta}{black}

\begin{document}

\title{EgoAVFlow: Robot Policy Learning with Active Vision from Human Egocentric Videos via 3D Flow}


\author{Daesol Cho$^{1}$, Youngseok Jang$^{2}$, Danfei Xu$^{1}$, and Sehoon Ha$^{1}$

\thanks{$^{1} $Daesol Cho, Danfei Xu, and Sehoon Ha are with the School of Interactive Computing, Georgia Institute of Technology, Georgia, USA
        {\tt\footnotesize \{dcho302, danfei, sehoonha\}@gatech.edu}}%
\thanks{$^{2} $Youngseok Jang is with the InnoCORE AI-Transformed Aerospace Research Center, KAIST, Daejeon, Republic of Korea
        {\tt\footnotesize duscjs59@gmail.com}}%
}



\maketitle

\begin{abstract}

Egocentric human videos provide a scalable source of manipulation demonstrations; however, deploying them on robots requires active viewpoint control to maintain task-critical visibility, which human viewpoint imitation often fails to provide due to human-specific priors. We propose EgoAVFlow, which learns manipulation and active vision from egocentric videos through a shared 3D flow representation that supports geometric visibility reasoning and transfers without robot demonstrations. EgoAVFlow uses diffusion models to predict robot actions, future 3D flow, and camera trajectories, and refines viewpoints at test time with reward-maximizing denoising under a visibility-aware reward computed from predicted motion and scene geometry. Real-world experiments under actively changing viewpoints show that EgoAVFlow consistently outperforms prior human-demo-based baselines, demonstrating effective visibility maintenance and robust manipulation without robot demonstrations. Project page: \url{https://dscho1234.github.io/egoavflow/}



\end{abstract}


\section{Introduction}

Learning robot policies from human demonstrations is an appealing alternative to large-scale robot data collection, especially as egocentric human videos provide abundant everyday manipulation behaviors \cite{damen2018scaling, grauman2022ego4d}. These videos contain rich task intent and interaction patterns, suggesting a scalable path toward imitation and representation learning. However, directly transferring human demonstrations to robots remains challenging: the robot must not only reproduce the action, but also perform \emph{active perception} to understand the scene. \textcolor{blue}{In real-world manipulation, robots frequently need to adjust their camera viewpoints to keep task-critical information in view \cite{bajcsy2018revisiting}. Even briefly losing the object or target could cascade into execution failures. In this work, we aim to enable policies to actively choose their viewpoints instead of passively accepting camera streams.}


A seemingly straightforward solution is to imitate the human viewpoint from egocentric demonstrations. Recent works even collect demonstrations via hardware-based teleoperation interfaces that record human head motion (and sometimes gaze) and train robots to reproduce these viewpoints \cite{chuang2025active, xiong2025vision, chuang2025look}. \textcolor{blue}{However, naively imitating a human viewpoint is often suboptimal for robots. Egocentric human videos are produced under strong human-specific priors that do not directly translate to robotic perception. For example, human uses frequent, rapid saccadic eye movements to gather information, which may not be optimal for a learned policy.} Moreover, humans exhibit behaviors such as the vestibulo-ocular reflex \cite{habra2017multimodal, roncone2016cartesian}, where the head moves while gaze stabilizes on the object; a robot camera does not need to replicate such head motions. These factors yield training signals that may be internally consistent for humans but visually unreliable for robots. Therefore, the goal should not be to mimic human camera motion, but to learn an \emph{independent} viewpoint adjustment strategy that maintains task-critical visibility while executing the task.

\textcolor{magenta}{However, learning an independent view policy raises a key question: what should the policy reason over to decide where to look? Visibility depends on the future 3D configuration of the manipulated object, the end-effector, and the surrounding scene geometry. Thus, planning camera motion requires a predictive representation that (i) captures task-relevant motion in 3D, (ii) is compatible with geometric visibility computation under candidate viewpoints, and (iii) is robust to view-dependent appearance, while also being (iv) embodiment-agnostic so that policies trained on human videos remain applicable when the agent's embodiment differs at deployment. Standard 2D visual features entangle appearance with viewpoint and embodiment, providing no direct way to forecast where the object will be in 3D for visibility scoring. This representation gap is a key consideration in coupling manipulation plans with visibility-aware planning from egocentric human videos.}


To address this, we introduce a shared 3D flow representation that serves as a common interface across embodiments and across policies, supporting joint learning of manipulation and viewpoint control. It directly encodes task-relevant 3D motion of scene elements across time while discarding view-dependent appearance, making it robust to viewpoint changes and suitable for zero-shot transfer without robot demonstrations. We construct this representation by unprojecting a pixel tracker's output into 3D flow \cite{karaev2025cotracker3} and mapping human hand pose estimates \cite{pavlakos2024reconstructing} to the robot end-effector space. 


Building on this representation, we propose a framework for a robot policy learning from \textbf{Ego}centric human videos with \textbf{A}ctive \textbf{V}ision via a 3D \textbf{Flow} representation (\textbf{EgoAVFlow}). EgoAVFlow consists of three 3D flow-based components: (i) a robot manipulation policy that predicts future robot actions, (ii) a flow generation model that predicts future 3D flow describing the object's motion, and (iii) a view policy that outputs future camera viewpoints. Crucially, we define a visibility-aware reward using the predicted 3D flow, the predicted robot actions, and the environment geometry, and perform test-time reward-maximizing denoising \cite{li2024derivative, uehara2024fine} to obtain viewpoints that maximize future visibility. The diffusion prior naturally captures the human demonstrator's head-motion distribution, while the reward-maximizing denoising allows the camera to deviate from the human viewpoint whenever visibility requires it. This yields two independently functioning capabilities: visibility-aware viewpoint adjustment and 3D-aware manipulation.

Through real-world experiments, we evaluate EgoAVFlow under actively changing viewpoints and compare it against human-demo-based baselines. The results show that EgoAVFlow consistently outperforms prior works, demonstrating effective visibility maintenance and robust manipulation capability. In summary, the contributions of this work are: 

\begin{itemize}
    \item We propose a shared 3D flow representation that bridges the human-robot embodiment gap and unifies manipulation with viewpoint control, without robot data.

    \item Building on this, we introduce a viewpoint adjustment strategy that explicitly optimizes the visibility, enabling exploratory camera motions that are decoupled from the human demonstrator while maintaining visibility.

    \item Under such actively changing viewpoints, EgoAVFlow significantly outperforms prior human-demo-based robot learning methods, highlighting its 3D-aware perception and viewpoint-robust manipulation.

\end{itemize}

\begin{figure*}[t]
    \centerline{\includegraphics[width=\linewidth]{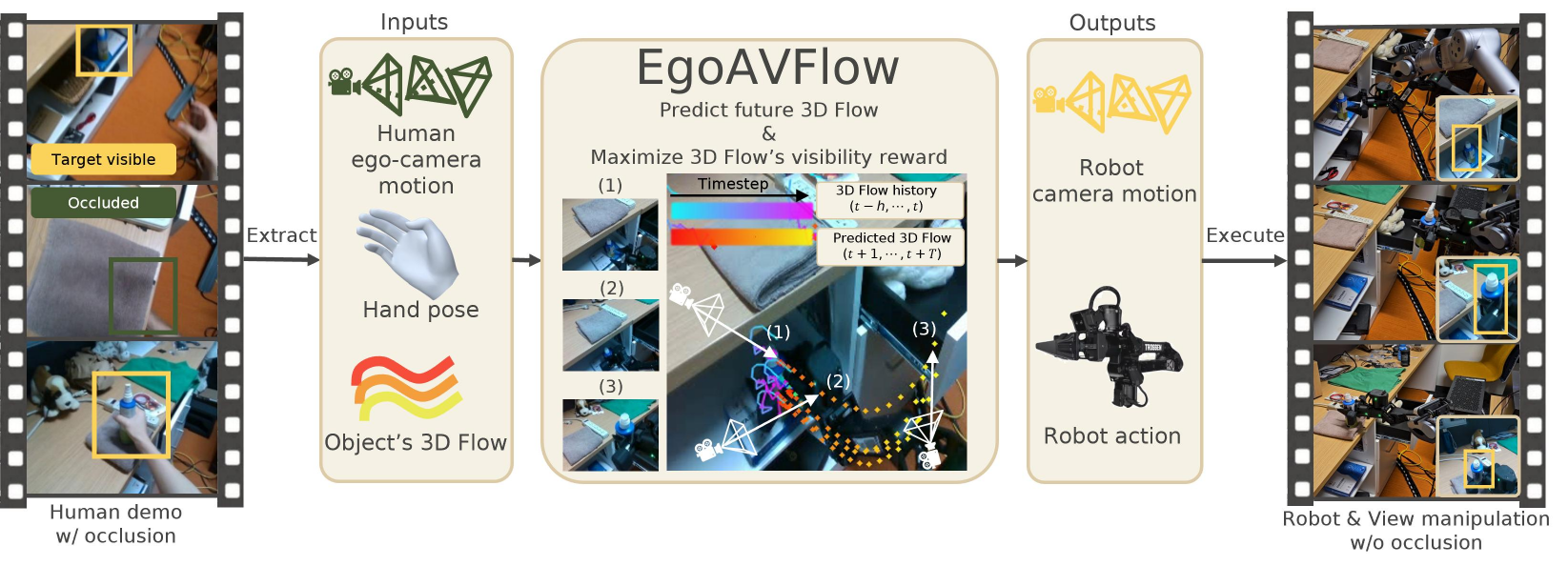}}
    \vspace{-0.2cm}
    \caption{\textcolor{magenta}{EgoAVFlow learns manipulation and active viewpoint control from egocentric human videos by predicting future 3D flow and optimizing camera viewpoints for visibility, yielding viewpoint-robust robot execution without robot demonstrations.}}

    \label{fig:thumbnail}
    \vspace{-0.6cm}
\end{figure*}

\section{Related Works}\


\noindent\textbf{\textcolor{magenta}{Learning from human demonstration}.} Building on recent progress in computer vision and robot learning, several approaches leverage human demonstration data to acquire robotic skills. Prior works either translate human videos into robot-centric observations via generative editing \cite{lepert2025phantom, smith2019avid}, infer manipulation affordances from human videos \cite{bahl2023affordances, chen2025vidbot}, or collect paired human-robot data through co-training pipelines and hardware setups (e.g., smart glasses) \cite{kareer2025egomimic, zhu2025emma, engel2023project}. Another line of work uses flow-based representations to mitigate the human-robot embodiment gap \cite{xu2024flow, collins2025amplify, haldar2025point, liu2025egozero}. Despite this progress on transfer across human-robot embodiments, these methods typically assume a passively set camera viewpoint and do not explicitly optimize viewpoint adaptation during execution. In contrast, we use 3D flow as a shared representation that \emph{both} bridges the embodiment gap and supports active viewpoint adjustment for visibility-aware manipulation.

\noindent\textbf{Active vision.} 
To address viewpoint variations, prior robotics works often cast next-best-view (NBV) selection as an active perception problem, targeting scene reconstruction \cite{xu2025area3d}, pose estimation \cite{wu2015active}, or uncertainty reduction \cite{zeng2020view}. For robotic manipulation, some approaches leverage novel-view synthesis to scale up training data \cite{chen2024rovi, tian2024view}, but they do not explicitly plan viewpoints during execution. Other recent works imitate human viewpoints using hardware-based robot teleoperation interfaces that track the operator's head motion to record egocentric camera trajectories \cite{xiong2025vision, chuang2025active, yu2025egomi}, optionally leveraging gaze information \cite{chuang2025look}; however, these methods largely assume that the human viewpoint strategy is optimal. Reinforcement learning (RL) has also been explored for viewpoint control \cite{shang2023active, wang2025observe, cheng2018reinforcement}, but such approaches typically require extensive on-policy interactions, making real-world training challenging. In contrast, we adapt viewpoints online via test-time reward-maximizing diffusion denoising: the human demonstrator's head motion provides a prior, while the camera pose is adapted to maximize visibility.

\section{Preliminary}\label{sec:preliminary}

\subsection{Data pre-processing}\label{subsec:data_preprocessing}


\paragraph{Robot data from egocentric human video}

We assume access to an egocentric human demonstration dataset $\mathcal{D}_{\text{human}}=\{\tau^i\}_{i=1}^L$ with $L$ video demonstrations, where each demonstration $\tau^i$ is a sequence of RGBD observations $\{I_t\}_{t=1}^{T'}$.
\textcolor{blue}{To derive robot-equivalent proprioception and actions from human videos, we estimate 3D hand keypoints and a 6DoF wrist pose using HaMeR \cite{pavlakos2024reconstructing}. Following prior work that maps egocentric hand motion to a gripper-equivalent interface \cite{lepert2025phantom}, we construct a gripper-equivalent 6DoF pose from the hand keypoints and infer a binary gripper command (open/close). Finally, we concatenate the gripper-equivalent position, orientation (6D rotation representation \cite{zhou2019continuity}), and the gripper command to form the robot proprioception $p_t \in \mathbb{R}^{10}$, and define the robot action as the next-step target in the same representation, $a_t \triangleq p_{t+1}$.}

\paragraph{Scene description via 3D flow}
To obtain motion cues over time, we track 2D pixels across frames using CoTracker3 \cite{karaev2025cotracker3}. This yields 2D pixel trajectories for $N$ query points, $\{\mathbf{u}_t\}_{t=1}^T \in \mathbb{R}^{N \times 3}$, where the first two channels correspond to image-plane coordinates $(x,y)$ and the last channel is a binary tracking indicator in $\{0,1\}$. Given depth at each tracked pixel, we unproject these 2D trajectories into 3D, resulting in $\{\mathbf{F}_t\}_{t=1}^T \in \mathbb{R}^{N \times 4}$, where the first three elements are the 3D point tracks in the camera coordinate frame and the last element is the tracking indicator. Since egocentric videos involve a moving camera, we also estimate the camera pose SE(3), $\mathbf{v}_t$, for each frame using DROID-SLAM \cite{teed2021droid} to interpret the 3D tracks and hand poses over time.

\paragraph{Marker coordinate representation}
Egocentric human videos exhibit diverse initial states, which leads SLAM to produce a different world coordinate frame for each demonstration. To express trajectories in a consistent reference frame, we convert all 3D quantities into a marker coordinate system defined by a ChArUco board. Specifically, robot actions and proprioception, 3D tracks, and camera poses are all represented in marker coordinates. For simplicity, we reuse the same notation $a_t$, $p_t$, $\mathbf{F}_t$, and $\mathbf{v}_t$ for their marker-frame counterparts in the remainder of this work.

\subsection{Soft Value-Based Denoising for Reward Maximizing Diffusion}\label{subsec:SVDD}

\textcolor{magenta}{A key component of our method is to refine diffusion-based predictions using a visibility-aware reward to guide the generation of future viewpoints at test time, so that samples improve reward while remaining consistent with the pre-trained diffusion prior. This subsection summarizes the core algorithmic primitive we use: soft value-based denoising. Compared to alternatives such as differentiable classifier-style guidance or RL fine-tuning, it does not require additional training and differentiable reward functions. This matches our setting, where the visibility reward is computed via geometric raycasting and must be applied at test time under changing scene reconstructions.}

\textcolor{magenta}{Assume the denoising dynamics of a pre-trained diffusion model are specified by a sequence of Markov kernels $\{p^{\mathrm{pre}}_{k-1}(\cdot|x_k)\}_{k=K}^{1}$ under the standard timestep convention $k=K,\ldots,1$. Let $p^{\mathrm{pre}}(\cdot)\in\Delta(\mathcal{X})$ denote the induced marginal distribution of the final sample $x_0$.} For notational simplicity, we suppress the conditioning context $c$ in notation; all distributions are understood to be conditional on $c$ when applicable. Given an arbitrary reward function $r(x_0)$, our goal is to bias generation toward high-reward samples while staying close to the pre-trained distribution.


\paragraph{Reward-tilted target distribution}
We consider the entropy-regularized objective
\begin{equation}
p^{(\alpha)}(\cdot)
\;\;=\;\;
\arg\max_{p \in \Delta(\mathcal{X})}
\; \mathbb{E}_{x \sim p}[r(x)]
\;-\;
\alpha \,D_{\mathrm{KL}}\!\bigl(p \,\|\, p^{\mathrm{pre}}\bigr),
\label{eq:entropy_reg_objective}
\end{equation}
whose solution corresponds to the reward-tilted distribution
\begin{equation}
p^{(\alpha)}(x)
\;\propto\;
\exp\!\bigl(r(x)/\alpha\bigr)\,p^{\mathrm{pre}}(x).
\label{eq:tilted_dist}
\end{equation}

Here, $\alpha>0$ controls the reward-naturalness trade-off, and $\alpha\!\to\!0$ yields a greedy
reward maximization behavior.

\paragraph{Soft value as a look-ahead score}


\textcolor{magenta}{Li et al. \cite{li2024derivative}} introduce a soft value function $v_{k-1}(\cdot)$ that measures how likely an intermediate noisy state
$x_{k-1}$ will lead to a high reward at the end of denoising:
\begin{equation}
v_{k-1}(x_{k-1})
\;:=\;
\alpha \log \mathbb{E}_{x_0 \sim p^{\mathrm{pre}}(\cdot \mid x_{k-1})}
\Bigl[\exp\!\bigl(r(x_0)/\alpha\bigr)\Bigr],
\label{eq:soft_value_def}
\end{equation}
where $\mathbb{E}_{x_0\sim p^{\mathrm{pre}}(\cdot\mid x_{k-1})}[\cdot]$ is a posterior-mean estimate \cite{song2020denoising}, induced by the pre-trained denoising dynamics $\{p^{\mathrm{pre}}_{k-1}(\cdot|x_k)\}_{k=K}^{1}$. Intuitively, $v_{k-1}$ is a one-step look-ahead score that rates intermediate states by their expected final reward under the pre-trained denoising dynamics.

\paragraph{Value-weighted denoising process}
Using $v_{k-1}$, we define a value-weighted denoising kernel
\begin{equation}
p^{\star,\alpha}_{k-1}(\cdot \mid x_k)
\;\propto\;
p^{\mathrm{pre}}_{k-1}(\cdot \mid x_k)\,
\exp\!\bigl(v_{k-1}(\cdot)/\alpha\bigr),
\label{eq:soft_opt_policy}
\end{equation}
which prefers candidates that are predicted to yield higher final reward.
Sequentially sampling with $\{p^{\star,\alpha}_{k-1}\}_{k=K}^1$ induces the target distribution in
Eq.~\eqref{eq:tilted_dist}, hence optimizing the entropy-regularized objective in
Eq.~\eqref{eq:entropy_reg_objective}.

\paragraph{Denoising via per-step importance resampling}
Direct sampling from Eq.~\eqref{eq:soft_opt_policy} is intractable in general due to the normalizer. Therefore, we approximate each step by: (i) drawing $M$ candidates from the proposal $p^{\mathrm{pre}}_{k-1}(\cdot \mid x_k)$, (ii) assigning weights $w \propto \exp(\hat v_{k-1}/\alpha)$ using an estimated value function $\hat v$, and (iii) selecting one candidate by categorical resampling, i.e.,

\begin{equation}
\resizebox{\columnwidth}{!}{$
p_{k-1}^{\star, \alpha}\left(\cdot \mid x_k\right) \approx
\sum_{m=1}^M \frac{w_{k-1}^{\langle m\rangle}}{\sum_{j=1}^M w_{k-1}^{\langle j\rangle}}
\delta_{x_{k-1}^{\langle m\rangle}},
\left\{x_{k-1}^{\langle m\rangle}\right\}_{m=1}^M \sim
p_{k-1}^{\mathrm{pre}}\left(\cdot \mid x_k\right),
$}
\label{eq:categorical_resampling}
\end{equation}
where $w_{k-1}^{\langle m\rangle}:=\exp \left(v_{k-1}\left(x_{k-1}^{\langle m\rangle}\right) / \alpha\right)$ and $\delta_a$ denote a Dirac delta distribution centered at $a$. This yields an inference-time optimization that approximately samples from the value-weighted denoising process and consequently maximizes the soft value along the denoising trajectory. We denote $\hat{x}_0(x_k)\approx \mathbb{E}_{x_0\sim p^{\mathrm{pre}}(\cdot\mid x_k)}[x_0]$ as a posterior-mean estimate \cite{song2020denoising}, where $p^{\mathrm{pre}}(\cdot\mid x_k)$ is the same as in Eq.~\eqref{eq:soft_value_def}. Then, we use $r(\hat{x}_0(x_k))$ as an estimation for $v_k(x_k)$ without additional training. More rigorous theoretical details of value-weighted denoising can be found in \cite{li2024derivative}.

\section{EgoAVFlow: Policy learning from egocentric human videos with active vision via a 3D flow}

\begin{figure*}[t]
    \centerline{\includegraphics[width=\linewidth]{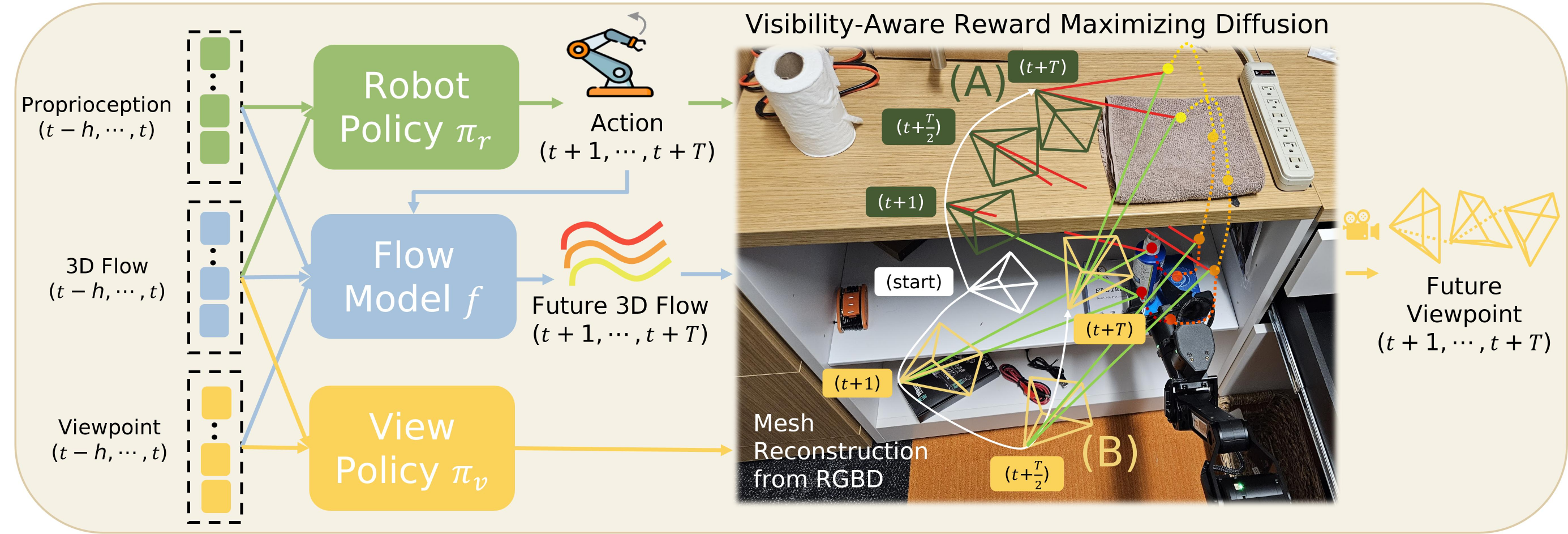}}
    \caption{\textbf{Method overview.} EgoAVFlow consists of three diffusion models. The robot policy $\pi_r$ produces future robot action sequences. The flow generation model $f$ predicts future 3D flows from the outputs of $\pi_r$. The view policy $\pi_v$ produces future camera viewpoints from the outputs of $\pi_r$, $f$, and reconstructed mesh surfaces through a visibility-aware reward-maximizing denoising process. \textcolor[HTML]{445A33}{\textbf{Viewpoints (A)}} represent that most query points are invisible (\textcolor[HTML]{E82A27}{\textbf{Red LOS}}) due to the table's mesh surface or out of FoV, whereas in \textcolor[HTML]{FBD35A}{\textbf{viewpoints (B)}} these points are visible (\textcolor[HTML]{92D050}{\textbf{Green LOS}}), yielding a higher visibility reward.} 
    \label{fig:method}
    \vspace{-0.6cm}
\end{figure*}

\subsection{Overall Framework}\label{method:three-stage-model} 


\textcolor{magenta}{Our goal is to learn a manipulation policy and an independent view adjustment strategy from egocentric human videos. This setting raises two coupled challenges. First, viewpoint control cannot be trained by straightforward imitation: the recorded human camera motion is not optimized for robotic visibility. Second, deciding where to move the camera requires reasoning about future scene motion and geometric occlusions, which depend on both the robot's planned interaction and the environment geometry.}

\textcolor{magenta}{These considerations motivate a modular design with three components: (i) a manipulation policy that proposes future robot actions, (ii) a predictive 3D flow representation that enables visibility reasoning under candidate viewpoints, and (iii) a view policy that provides a strong prior over plausible camera motions while allowing test-time optimization under a visibility-aware reward.} 

We implement all three components with diffusion models by using preprocessed data in Section \ref{sec:preliminary}. For these three models, we adopt a diffusion transformer-based backbone \cite{liu2024rdt} that injects condition tokens via cross-attention, train these models using a standard DDPM \cite{ho2020denoising} framework, and obtain samples using DDIM \cite{song2020denoising}. We set the prediction horizon $T=24$ for all three models, and iteratively sample action chunks for every $H=12$ steps, using the first 12 elements from the action chunks. Fig.~\ref{fig:method} shows an overview of the proposed framework.


\paragraph{Robot policy $\pi_r$}
The robot policy $\pi_r$ is trained to output robot action chunks $\hat{a}_{t+1:t+T}$ from the 3D flow tracking history $\mathbf{F}_{t-h:t}$, proprioception history $p_{t-h:t}$, where $h$ is history length. 

\paragraph{Future flow generation model $f$}
The flow generation model $f$ is trained to predict future 3D flows \textcolor{blue}{$\hat{\mathbf{F}}_{t+1:t+T} \in \mathbb{R}^{N\times T \times 4}$.}
It takes as input the same context as $\pi_r$, and is additionally conditioned on the viewpoint history
$\mathbf{v}_{t-h:t}$ and the future robot action sequence $\hat{a}_{t+1:t+T}$ predicted by $\pi_r$.
This is because future 3D flow depends on both camera motion and robot motion. This model captures how the robot's planned interaction (via $\hat{a}_{t+1:t+T}$) induces future scene changes, providing a compact intermediate representation for ``what will move where''. These predicted future flows \textcolor{blue}{$\hat{\mathbf{F}}_{t+1:t+T}$ are used as query points for visibility computation} in Section \ref{subsec:reward}. 


\paragraph{View policy $\pi_v$}
The view policy $\pi_v$ is trained to predict future camera viewpoints $\hat{\mathbf{v}}_{t+1:t+T}$. It takes as input the same context as $f$, except for the proprioception history. At inference time, we do not simply sample $\hat{\mathbf{v}}_{t+1:t+T}$ from $\pi_v$; instead, we apply soft value-based denoising to bias generation toward viewpoints that are optimal under the visibility-aware reward in Section~\ref{subsec:reward}. The overall algorithm is summarized in Algorithm~\ref{alg:overview}.


\subsection{View Selection via Visibility-Aware Reward}\label{subsec:reward}



\textcolor{magenta}{We formulate view selection as choosing a future camera trajectory $\hat{\mathbf{v}}_{t+1:t+T}$ that keeps task-relevant scene elements observable during the robot's planned interaction. Scoring candidate views requires predicting future 3D configurations and checking occlusions / FoV via mesh raycasting, which yields a non-differentiable objective. Therefore, we treat $\pi_v$ as a learned prior over plausible camera motions and refine its samples at inference time using a visibility-aware reward through reward-guided denoising (Sec.~\ref{subsec:SVDD}). This preserves the diffusion prior while allowing deviations when visibility demands it.}

\textcolor{magenta}{Our reward prioritizes the visibility of query points defined from predicted future 3D flows: $\{\mathbf{q}_{t,i}\in\mathbb{R}^3\}_{t=t+1:t+T}^{i=1:N}$ are obtained by dropping the indicator channel of $\hat{\mathbf{F}}_{t+1:t+T}$. We evaluate visibility by raycasting from candidate camera poses to $\{\mathbf{q}_{t,i}\}$ against the union of an environment mesh $\mathcal{M}^e$ (reconstructed up to time $t$ with Nvblox \cite{millane2024nvblox}) and a time-varying robot mesh $\mathcal{M}^r_t$ (constructed from $\hat{a}_{t+1:t+T}$ via inverse kinematics). The resulting visibility term defines $R_{\mathrm{vis}}$, which forms the core of our reward; we then augment it with auxiliary terms for stability and safety.}

\subsubsection{Visibility reward}
\label{subsec:visibility_reward}
In this work, we define visibility as the conjunction of (a) whether there exist any obstacles between the camera origin and query points, and (b) whether the query point is within the field-of-view (FoV) of the camera. For (a), we define the line-of-sight (LOS) segment from the camera center for each query point $\mathbf{q}_{t,i}$:

\begin{equation}
\ell_{t,i} \;=\; \left\{\mathbf{v}_{t,\mathrm{pos}} + \lambda(\mathbf{q}_{t,i}-\mathbf{v}_{t,\mathrm{pos}})\;\middle|\;\lambda\in[0,1]\right\},
\end{equation}
where \textcolor{blue}{$\mathbf{v}_{t,\mathrm{pos}} \in \mathbb{R}^3$} is the position component from the camera pose prediction $\hat{\mathbf{v}}_t$. Then, we define an unobstructed-LOS indicator using mesh intersection:


\begin{equation}
s_{t,i} \;=\; \mathbb{I}\!\left[\ell_{t,i}\cap\mathcal{M}_t=\emptyset\right], \quad \mathcal{M}_t=\mathcal{M}^e \cup \mathcal{M}^r_t,
\end{equation}
where mesh intersection is computed by raycasting. \textcolor{blue}{Intuitively, $s_{t,i}=1$ indicates there is nothing between the camera and the query point, while $s_{t,i}=0$ indicates the query point is occluded by the mesh surfaces.} For (b), we project the query point to the image plane using the camera projection $\Pi_t(\cdot)$:

\begin{equation}
(u_{t,i},v_{t,i}) \;=\; \Pi_t(\mathbf{q}_{t,i}),
\end{equation}
where $(u_{t,i},v_{t,i})$ are pixel coordinates and the image size is $W_{img}\times H_{img}$. We define an in-FoV indicator:

\begin{equation}
f_{t,i} \;=\; \mathbb{I}\!\left[0 \le u_{t,i} < W_{img} \;\wedge\; 0 \le v_{t,i} < H_{img}\right].
\end{equation}

The binary visibility reward for point $\mathbf{q}_{t,i}$ is the conjunction, and we average over time and query points:

\begin{equation}
R_{\mathrm{vis}}
\;=\;
\frac{1}{TN}\sum_{t=t+1}^{t+T}\sum_{i=1}^{N} r^{\mathrm{vis}}_{t,i}, \quad r^{\mathrm{vis}}_{t,i}=s_{t,i} f_{t,i}
\label{eq:r_vis_final}.
\end{equation}

A visual illustration of computing $R_{\mathrm{vis}}$ is shown in Fig. \ref{fig:method}.

\begin{algorithm}[t]
\caption{EgoAVFlow}
\label{alg:overview}
\begin{algorithmic}[1]
\STATE {\bfseries Require:}
$\pi_r$, $\pi_v$, $f$, robot\_env, view\_env, $t\!\leftarrow\!0$,
chunk size $H(\leq T)$, horizon $T$,
environment mesh $\mathcal{M}^e$

\WHILE{not done}
    \IF{$t \bmod H = 0$}
        \STATE $\hat{a}_{t+1:t+T} \sim \pi_r(\cdot|p_{t-h:t}, \mathbf{F}_{t-h:t})$ 
        \STATE $\hat{\mathbf{F}}_{t+1:t+T} \sim f(\cdot|p_{t-h:t}, \mathbf{F}_{t-h:t}, \mathbf{v}_{t-h:t}, \hat{a}_{t+1:t+T})$ 
        \STATE $\hat{\mathbf{v}}_{t+1:t+T} \sim$ \\ RewMaxDiff $(\pi_v, \hat{\mathbf{F}}_{t+1:t+T}, \hat{a}_{t+1:t+T}, \mathcal{M}^e)$.
        \STATE Get $p_{t+1:t+H}$, $\mathbf{v}_{t+1:t+H}$, $\mathbf{F}_{t+1:t+H}$ from \\ robot\_env.step($\hat{a}_{t+1:t+H}$), view\_env.step($\hat{\mathbf{v}}_{t+1:t+H}$)
        \STATE $t \leftarrow t + H$
    \ENDIF
\ENDWHILE

\vspace{0.2em}
\STATE {\bfseries Func} RewMaxDiff $\bigl(p^{\mathrm{pre}}(\cdot), \hat{\mathbf{F}}_{t+1:t+T}, \hat{a}_{t+1:t+T}, \mathcal{M}^e \big)$:
\STATE \hspace{1em} \textbf{for} $k = K, \ldots, 1$ \textbf{do}
\STATE \hspace{2em} Sample $\{x_{k-1}^{(i)}\}_{i=1}^{M} \sim p^{\mathrm{pre}}_{k-1}(\cdot \mid x_k)$

\STATE \hspace{2em} \textbf{for} $i = 1, \ldots, M$ \textbf{do}
\STATE \hspace{3em} $v_{k-1}^{(i)} \leftarrow R\!\Big(\hat{x}_0(x_{k-1}^{(i)}), \hat{\mathbf{F}}_{t+1:t+T}, \hat{a}_{t+1:t+T}, \mathcal{M}^e\Big)$ \hfill (Eq.~\eqref{eq:r_total})
\STATE \hspace{3em} $w_{k-1}^{(i)} \leftarrow \exp\!\big(v_{k-1}^{(i)}/\alpha\big)$
\STATE \hspace{2em} \textbf{end for}

\STATE \hspace{2em} $x_{k-1} \gets x_{k-1}^{(i^\star)}$, where
$i^\star \sim \mathrm{Cat}\Bigl(\dfrac{w_{k-1}^{(i)}}{\sum_{j=1}^{M} w_{k-1}^{(j)}}\Bigr)$
\STATE \hspace{1em} \textbf{end for}
\STATE \hspace{1em} \textbf{Output:} $x_0$
\STATE {\bfseries End Func}

\end{algorithmic}
\end{algorithm}

\subsubsection{Auxiliary reward terms}\label{subsubsec:aux_reward}
Since $r^{\mathrm{vis}}_{t,i}$ is binary, it lacks a continuous objective that reflects the quality of the current viewpoint and provides an informative optimization signal. In addition, multiple viewpoints can satisfy full visibility. Thus, we additionally add a weighted sum of the following auxiliary terms:

\paragraph{Close to query points}
We prefer \textcolor{blue}{camera positions} not excessively far from the query points:

\begin{equation}
R_{\mathrm{close}}=\frac{1}{TN}\sum_{t=t+1}^{t+T}\sum_{i=1}^{N}\exp{(-\left\lVert \mathbf{v}_{t,\mathrm{pos}} - \mathbf{q}_{t,i}\right\rVert_2)}.
\end{equation}

\paragraph{Camera margin}
To discourage viewpoints where query points are only barely visible and instead favor views that keep them visible under small pose perturbations, we generate $J$ perturbed viewpoints $\{\tilde{\mathbf{v}}_t^{(j)}\}_{j=1}^{J}$ from $\mathbf{v}_t$ by adding Gaussian noise to translation and rotation, i.e., $\tilde{\mathbf{v}}_{t,\mathrm{pos}}^{(j)} = \mathbf{v}_{t,\mathrm{pos}} + \epsilon_{\mathrm{pos}}^{(j)}$ with $\epsilon_{\mathrm{pos}}^{(j)} \sim \mathcal{N}(\mathbf{0}, \sigma_{\mathrm{pos}}^2\mathbf{I})$ and $\sigma_{\mathrm{pos}}=0.02\,\mathrm{m}$, and similarly for rotation with $\sigma_{\mathrm{rot}}=0.05\,\mathrm{rad}$. For each perturbation, we compute $R_{\mathrm{vis}}^{(j)}$ (Eq.~\eqref{eq:r_vis_final}) and define

\begin{equation}
R_{\mathrm{marg}}
=
\min_{j \in \{1,\dots,J\}} R_{\mathrm{vis}}^{(j)}
\;\cdot\;
\Bigl(1 - \lambda_{\mathrm{var}}\cdot \mathrm{Var}(\{R_{\mathrm{vis}}^{(j)}\}_{j=1}^{J})\Bigr),
\label{eq:r_margin}
\end{equation}
where $\lambda_{\mathrm{var}}=0.1$. 



\paragraph{Safety}
To penalize camera viewpoints that are too close to the robot end effector, we define a safety reward $R_{\mathrm{safe}}$ based on the distance between the predicted camera position $\mathbf{v}_{t,\mathrm{pos}}$ and the predicted end-effector position $\hat{a}_t$ at each horizon step:

\begin{equation}
R_{\mathrm{safe}}
=
-\frac{1}{T}\sum_{t=t+1}^{t+T}
\exp\!\left(-\frac{\|\mathbf{v}_{t,\mathrm{pos}}-\hat{a}_t\|_2}{\sigma_{\mathrm{safe}}}\right),
\quad
\sigma_{\mathrm{safe}}=0.1\,\mathrm{m},
\label{eq:r_safe}
\end{equation}
where \textcolor{blue}{$\sigma_{\mathrm{safe}}=0.1\,\mathrm{m}$}. Then, the following reward composition is used for \textcolor{blue}{the RewMaxDiff function in Algorithm \ref{alg:overview}}:

\begin{equation}
\begin{aligned}
R(\hat{\mathbf{v}}_{t+1:t+T}, \hat{\mathbf{F}}_{t+1:t+T}, \hat{a}_{t+1:t+T}, \mathcal{M}^e) &\triangleq \\ 
R_{\mathrm{vis}}
+ \lambda_{\mathrm{c}} R_{\mathrm{close}}
+ \lambda_{\mathrm{m}} R_{\mathrm{marg}} &
+ \lambda_{\mathrm{s}} R_{\mathrm{safe}},
\end{aligned}
\label{eq:r_total}
\end{equation}
where $\lambda_{\mathrm{c}}, \lambda_{\mathrm{m}},  \lambda_{\mathrm{s}}$ are scalar weighting coefficients. 


\section{Experiments}\label{sec:experiment}

\textcolor{magenta}{In this section, we evaluate EgoAVFlow against baselines to support three findings: (i) fixed viewpoints cannot reliably maintain visibility during manipulation, (ii) directly imitating human viewpoints is insufficient for visibility-aware viewpoint adjustments, and (iii) conditioning policies on 3D flow yields the strongest performance under actively varying viewpoints.}

\begin{figure}[t]
    \centerline{\includegraphics[width=\linewidth]{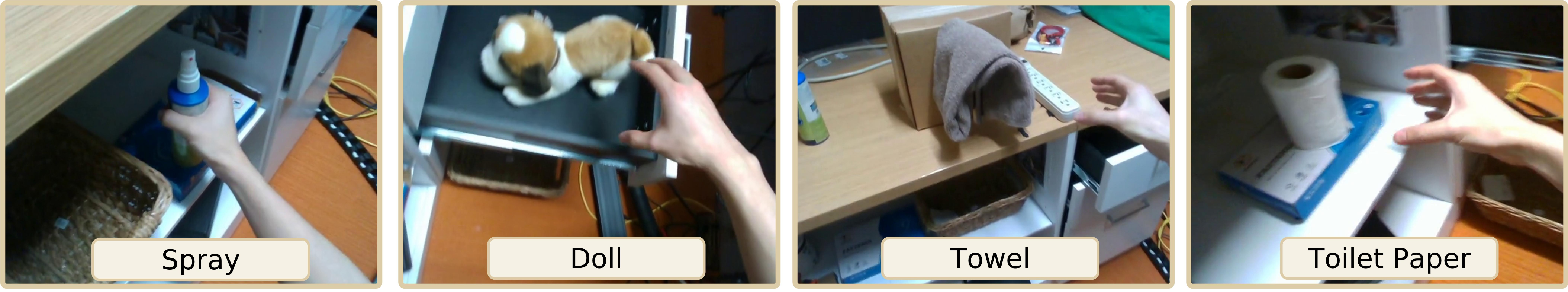}}
    \caption{\textbf{Tasks.} Each task requires appropriate viewpoint adjustments. Otherwise, the object is occluded by the robot or elements in the environment, such as a table or drawer.} 
    \label{fig:dataset}
    \vspace{-0.6cm}
\end{figure}

As an evaluation benchmark, we set up 4 tasks (Fig. \ref{fig:dataset}), where the camera viewpoint should be adjusted during rollout to maintain a fully visible status. For the dataset, we collect 150 egocentric human videos for each task by using a head-mounted RealSense D435, RGBD camera. \textcolor{blue}{For the query points, we annotate six points ($N=6$) on the object and goal location, such as a basket, though the method is agnostic to the number of points and scales naturally.} For hardware setup, we use the Trossen WidowX robot for manipulation (robot\_env), and the Unitree Z1 robot with a D435 camera mount for viewpoint adjustment (view\_env). Since the outputs of $\pi_r$ and $\pi_v$ are each robot's end-effector pose, we solve inverse kinematics to compute the target joint positions and control both robots at 4Hz to reach the target joints.


\subsection{\textcolor{magenta}{Continuous viewpoint adjustment is necessary for reliable visibility}}\label{subsec:Q1}
\begin{figure}[t]
    \centerline{\includegraphics[width=\linewidth]{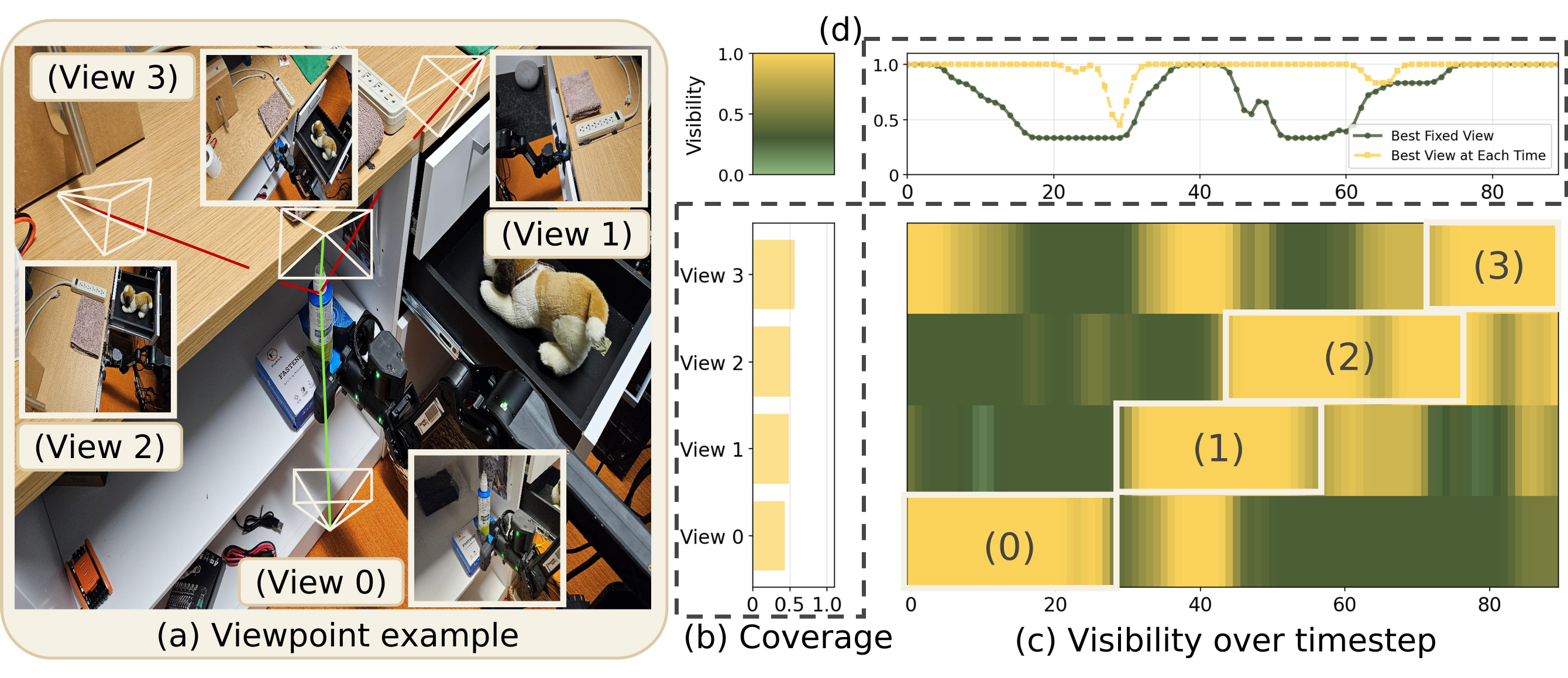}}
    \caption{\textbf{Visibility comparison (best viewed in the digital version).} The visibility is computed from each different fixed viewpoint. No single viewpoint can maintain full visibility throughout the execution, indicating that the viewpoint must be continuously adjusted online to maximize visibility.} 
    \label{fig:visibility_comparison}
    \vspace{-0.6cm}
\end{figure}

For an intuitive understanding of the necessity of viewpoint adjustment, we set up four fixed cameras for the spray task (\textcolor{blue}{Fig. \ref{fig:visibility_comparison}-(a)}) and computed the visibility for all timesteps in a demonstration video. Denoting $k_{v,t} = \frac{\text{number of visible query points at view } v \text{ and time } t}{\text{total number of query points}}$, we compute a coverage, $C_v = \frac{1}{T} \sum_{t=1}^{T} \mathbf{1}[k_{v,t} \geq 0.7]$. Then, we sort the view indices in descending order based on the coverage \textcolor{blue}{(Fig. \ref{fig:visibility_comparison}-(b))}, and represent high $k_{v,t}$ as yellow color, and low $k_{v,t}$ as dark-green color \textcolor{blue}{(Fig. \ref{fig:visibility_comparison}-(c))}. 

The figure shows that, despite nearly similar coverage among the viewpoints, no viewpoint can maintain full visibility during the rollout, and the best visible viewpoint also keeps changing (e.g., view index $0\rightarrow1\rightarrow2\rightarrow3$). It is further elaborated in the upper right figure \textcolor{blue}{(Fig. \ref{fig:visibility_comparison}-(d))}, which represents that even a fixed view with the highest coverage (\textcolor[HTML]{445A33}{dark-green line}) cannot provide the best visibility (\textcolor[HTML]{FBD35A}{yellow line}) during the episode rollout. This indicates that in most cases, the camera viewpoint should be adjusted to provide informative scene observations to the robot manipulation policy. 


\subsection{\textcolor{magenta}{Visibility-aware viewpoint planning outperforms human viewpoint imitation}}\label{subsec:Q2}
If the viewpoint adjustment is required, the next question is which viewpoints to follow. More specifically, we validate the recent approach of imitating human viewpoints \cite{chuang2025active, xiong2025vision}. Because both of these prior works utilize a robot teleoperation setting with a head-mounted VR device, they directly imitate the camera-mounted robot's joint angle or the end-effector's pose from image inputs. However, we do not have access to such information since we only have access to the egocentric human video. Therefore, the conditional diffusion model $\pi_v$ without the soft value-based denoising process is used as an implementation for the viewpoint imitation baseline, since it is analogous to the direct imitation of the camera viewpoints from corresponding inputs, and conceptually the same as the prior works. \textcolor{blue}{To isolate the effect of viewpoint imitation, we use the same manipulation policy as our method (i.e., $\pi_r$). We refer to it as a Human Viewpoint Imitation (\textbf{HVI})}. 


For comparison, we evaluate EgoAVFlow and \textcolor{blue}{HVI} over 25 trials per task and report the success rates. A rollout is considered successful if the robot manipulates the target object as desired \emph{and} the object remains visible throughout the episode. If the object drifts out of FoV, we treat this trial as a failure. 

\begin{figure}[h]
\centerline{\includegraphics[width=\linewidth]{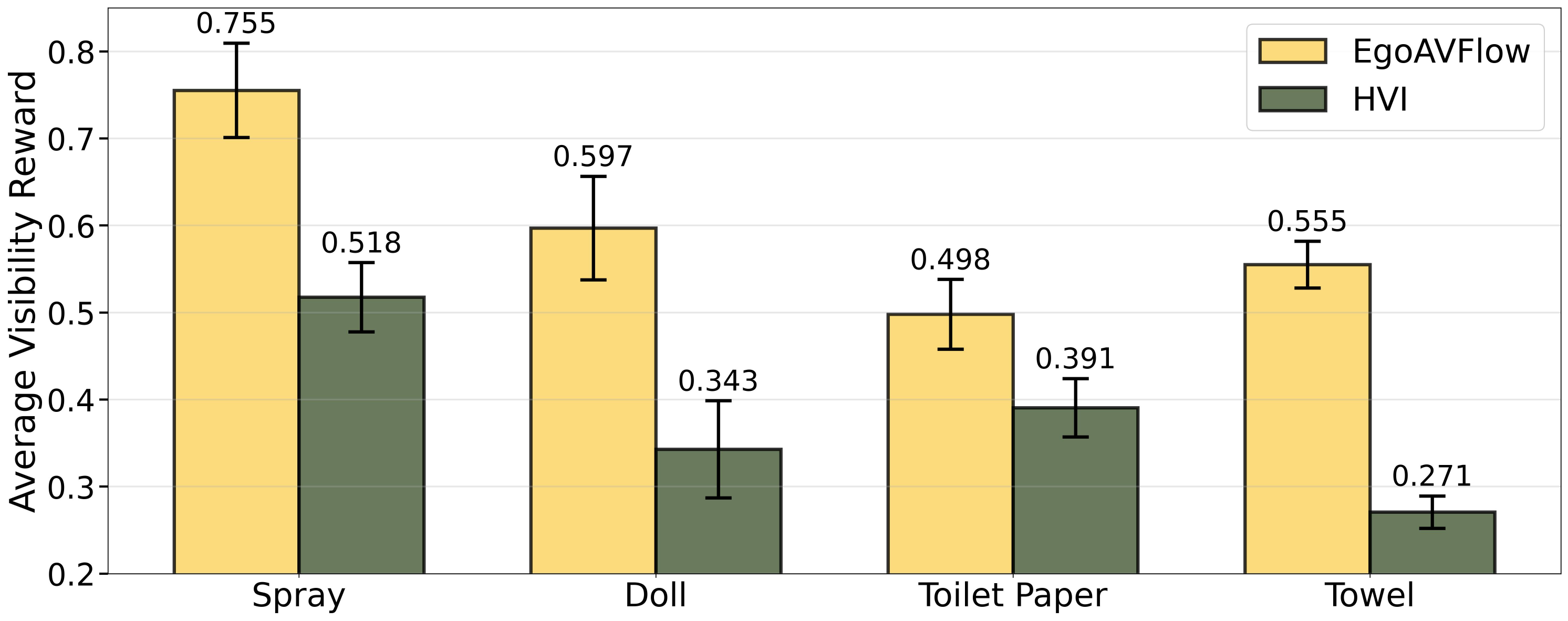}}
\caption{\textbf{Visibility reward.} For all tasks, EgoAVFlow achieves higher average visibility rewards $R_{vis}$ than \textcolor{blue}{HVI}, demonstrating our method's visibility maintenance capability. The error bars represent 1 standard error.}
\label{fig:reward}
\vspace{-0.6cm}
\end{figure}

\begin{figure*}[t]
    \centerline{\includegraphics[width=\linewidth]{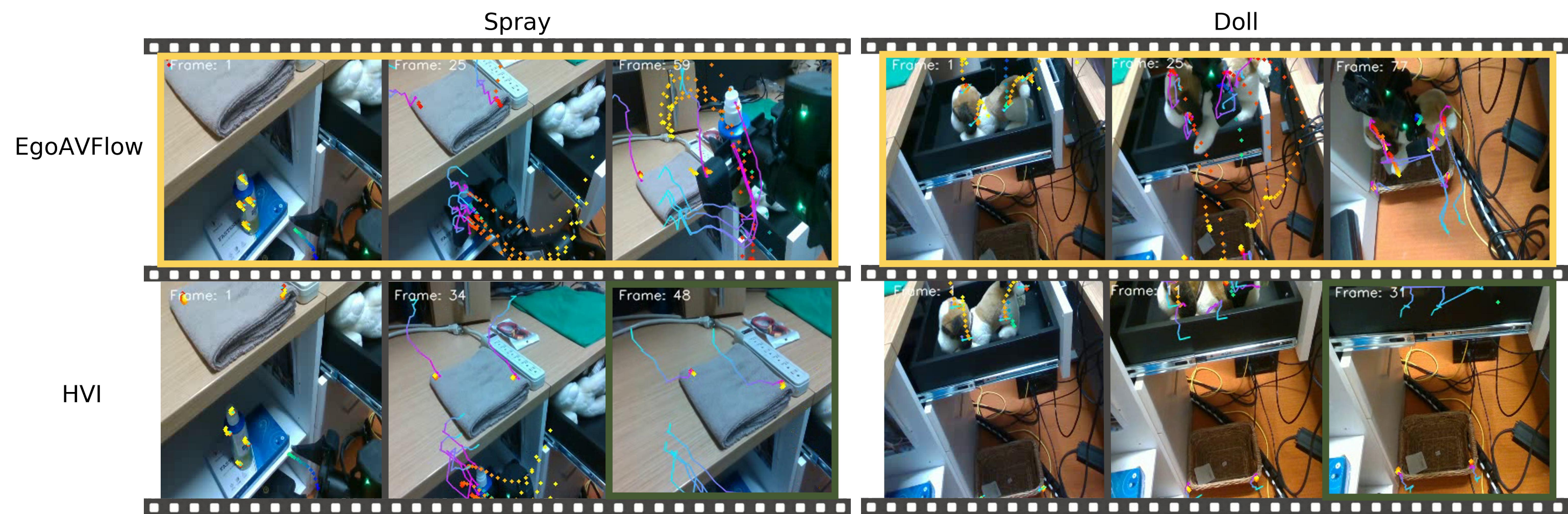}}
    \caption{\textbf{Qualitative comparison.} Due to the visibility-maximizing viewpoint adjustments, EgoAVFlow maintains visibility of the query points and their predicted future flows, whereas \textcolor{blue}{HVI} fails to keep them in view, causing the query points to move out of the FoV. All experimental figures and videos in Section \ref{sec:experiment} are best viewed in the supplementary video.} 
    \label{fig:qualitative}
    \vspace{-0.4cm}
\end{figure*}

As shown in Table \ref{table:policy}, EgoAVFlow outperforms \textcolor{blue}{HVI} in terms of success rates. Qualitatively, as shown in Fig. \ref{fig:qualitative}, EgoAVFlow maintains reliable visibility of the query points during rollout. These results are consistent with the quantitative visibility analysis in Fig. \ref{fig:reward}. For all tasks, EgoAVFlow achieves a higher visibility reward $R_{vis}$, \textcolor{blue}{which is consistent with the observed success rates, suggesting that maintaining visibility is important for task completion.} All of these results are attributed to the proposed visibility-maximizing diffusion for viewpoint planning. 

On the contrary, \textcolor{blue}{HVI} often fails as it tries to imitate the human viewpoints that do not consider occlusion, or near/outside the FoV, or produce weird values when faced with out-of-distribution viewpoint inputs, attributed to the inherent embodiment gap between humans' heads and robots' workspace limit. This demonstrates that it is crucial to have a capability that can actively adjust viewpoints for our preferences, rather than closely following the viewpoints in the dataset.

\subsection{\textcolor{magenta}{3D flow policy outperforms under actively varying viewpoints}}\label{subsec:Q3}
Now, given the non-human-imitating viewpoint adjustment strategy, we next ask which representation is effective for the robot policy's robust capability under such actively varying viewpoints, while using only human data. For this question, we compare our method with the following representative robot policy learning methods, designed to leverage human data. \textbf{AMPLIFY}\cite{collins2025amplify}: A 2D flow-based method that quantizes 2D flow tracking into a sequence of discrete codebooks, and learn a policy conditioned on these codebooks and an image input. \textbf{EgoZero}\cite{liu2025egozero}: A 3D flow-based method, similarly designed to our method. However, it depends on the triangulation under a static-object assumption to compute 3D flow, which can fail when the object is not static. \textbf{Phantom}\cite{lepert2025phantom}: A method that removes humans from egocentric videos via diffusion-based inpainting and overlays a robot onto the resulting frames, enabling policy training on synthesized robot observations. 

All baselines are trained on the same dataset as our method, and no real robot data is used. As AMPLIFY's policy also requires an image input, we provide it with the same synthesized robot image data used by Phantom. To isolate the effect of the policy representation, we use the same viewpoint adjustment module for all methods: our view policy $\pi_v$ with visibility-maximizing denoising. 

We evaluate the success rates using the same criteria as in Section \ref{subsec:Q2}, and the results are reported in Table \ref{table:policy}. EgoAVFlow shows superior capability compared to baselines. \textcolor{blue}{Across all tasks, this corresponds to a \textbf{1.8-2.5$\times$} improvement over the second-best baseline.} As the viewpoint keeps changing during the evaluation by our proposed view policy, the results demonstrate that the proposed 3D flow-based policy is view-invariant and benefits from its inherent 3D representation, yielding a viewpoint-robust manipulation capability.


\begin{table}[t]
\centering
\resizebox{\linewidth}{!}{
\begin{tabular}{cccccc}
\hline
      & Method       & Spray & Doll & Toilet Paper & Towel \\ \hline
\ref{subsec:Q2}    & HVI    & 5/25  & 10/25 & 7/25         & 5/25  \\ \hline
\multirow{3}{*}{\ref{subsec:Q3}}
      & AMPLIFY & 4/25  & 6/25  & 9/25         & 9/25  \\
      & EgoZero & 10/25 & 7/25  & 7/25         & 9/25  \\
      & Phantom & 5/25  & 7/25  & 6/25         & 8/25  \\

\rowcolor[HTML]{FBD35A}
\cellcolor[HTML]{FBD35A}{} & \textbf{EgoAVFlow} & \textbf{20/25} & \textbf{18/25} & \textbf{17/25} & \textbf{18/25} \\
\rowcolor[HTML]{FBD35A}
\cellcolor[HTML]{FBD35A}{} &                    & {\scriptsize($\times\,2.0$)} & {\scriptsize($\times\,2.5$)}
                           & {\scriptsize($\times\,1.8$)} & {\scriptsize($\times\,2.0$)} \\
\hline
\end{tabular}
}
\caption{\textbf{Success rates of EgoAVFlow, HVI (Human Viewpoint Imitation), and robot policy learning baselines.} All methods are trained on the same egocentric human video dataset (no robot data). \textcolor{blue}{Relative improvements compared to the second-best robot policy learning baseline are shown in parentheses}.}
\label{table:policy}
\vspace{-0.6cm}
\end{table}

\subsection{Failure analysis}

\begin{figure}[t]
    \centerline{\includegraphics[width=\linewidth]{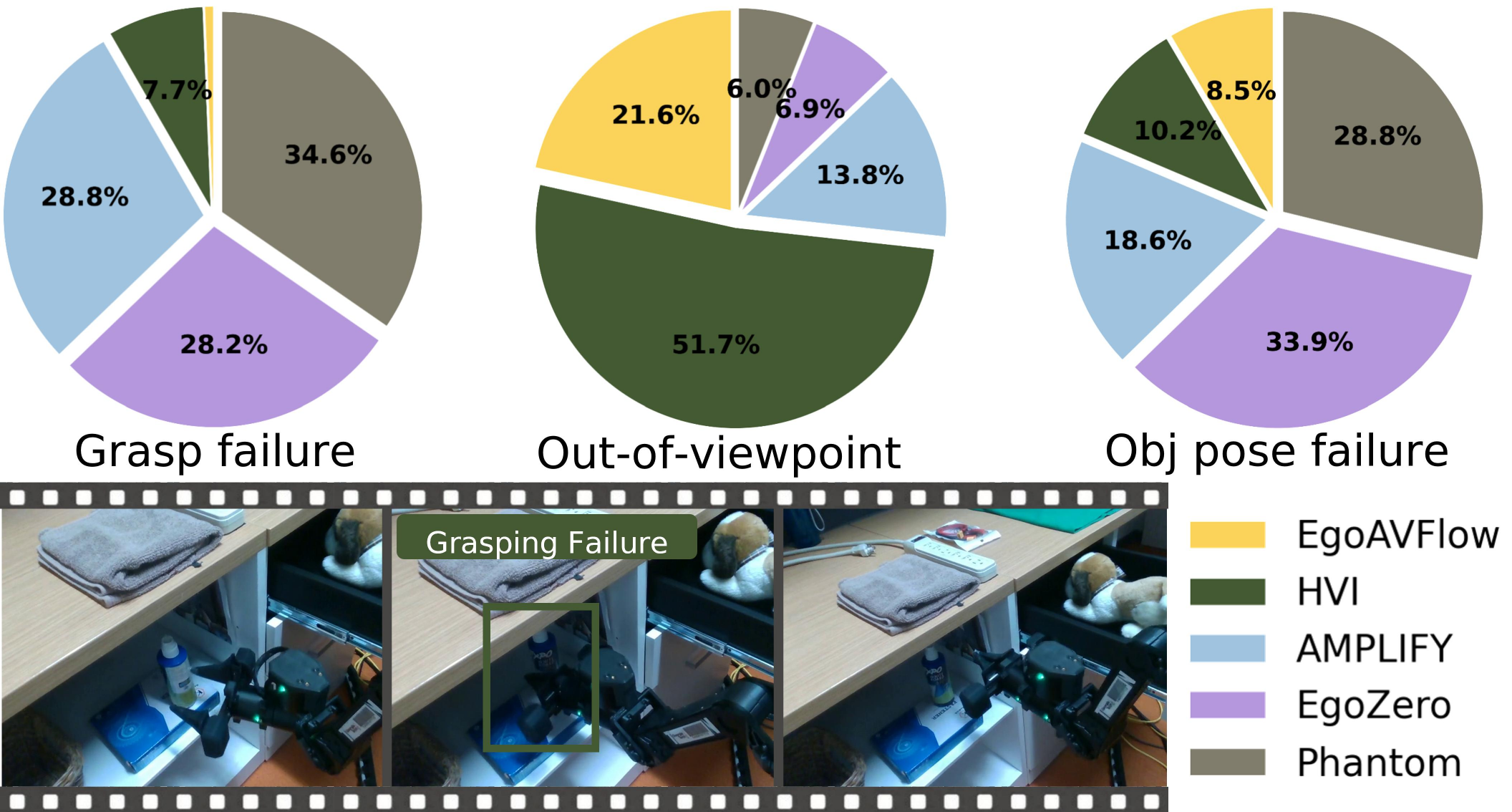}}
    \caption{\textbf{Failure analysis.} \textcolor{blue}{\emph{Top:} Method-wise composition of failures for each category. \emph{Bottom:} Example frames of a grasping failure. The breakdown suggests that manipulation-related failures in robot policy baselines are driven by a mismatch between their learned representations/assumptions, whereas human viewpoint imitation fails primarily due to the lack of visibility-aware viewpoint optimization.}}
    \label{fig:failure}
    \vspace{-0.6cm}
\end{figure}

We analyze failure cases to better understand the experiments for Section \ref{subsec:Q2}, \ref{subsec:Q3}. Specifically, we report the per-category breakdown of failures across methods, with the categories ordered chronologically. \textbf{Grasping failure}: The robot fails to grasp the object in the initial phase. \textbf{Out-of-viewpoint}: The robot succeeds in grasping, but the pixel tracker's tracking is lost, or the viewpoint is adjusted toward the wrong direction due to the inaccurate future flow prediction $\hat{\mathbf{F}}$. \textbf{Object pose failure}: Grasping and tracking succeed, but the robot fails to place the object in the demonstrated pose. 

The results are shown in Fig. \ref{fig:failure}. Since AMPLIFY and Phantom are not inherently 3D-aware, they suffer from distribution shifts when evaluated under the unseen viewpoints. EgoZero uses 3D points, but it cannot address the non-static scene, such as when the object moves due to the gripper's contact. As a result, these \textcolor{blue}{three robot policy baselines account for most manipulation-related failures: AMPLIFY, EgoZero, and Phantom together constitute 91.6\% in grasping failure, 81.3\% in object pose failure, and EgoAVFlow constitutes the smallest share, indicating stronger robustness in manipulation under actively varying viewpoints.} 

In the case of out-of-viewpoint, \textcolor{blue}{HVI} accounts for more than \textcolor{blue}{50\% of the failures} because it does not use our proposed view policy for visibility maintenance. EgoAVFlow accounts for the second-largest share of the failures. However, this does not indicate worse performance; rather, many robot policy baseline rollouts are already counted as early grasping failures and thus do not reach the later stages. If they progressed further, their proportions would be more comparable to our method.


\section{Conclusion}

In this work, we introduced a shared 3D flow-based representation that enables (i) joint control of the robot and camera viewpoint and (ii) mitigation of the human-robot embodiment gap when training solely from human data. We propose a visibility-maximizing robot manipulation pipeline, which consists of a 3D flow-based robot policy, a future 3D flow generation model, and a viewpoint adjustment policy. Across experiments, we demonstrate that EgoAVFlow outperforms prior works that try to follow the human's viewpoint in the dataset and achieves superior manipulation performance under actively varying viewpoints, driven by visibility maximization. Despite these gains, the current formulation assumes that the tracking points are observable at the initial timestep. In other words, our method does not address searching or reasoning about which points-of-interest should be tracked. Incorporating this capability is a promising direction for future work toward more autonomous robot manipulation.

\bibliographystyle{IEEEtran}
\bibliography{ref}

\vfill

\end{document}